\documentclass[letterpaper, 10 pt, conference]{ieeeconf}  % Comment this line out if you need a4paper

\IEEEoverridecommandlockouts                              % This command is only needed if 
\usepackage{amsmath}  % assumes amsmath package installed
\usepackage{amssymb}  % assumes amsmath package installed
\usepackage{siunitx}
\usepackage{subcaption}
\usepackage{verbatim}
\usepackage{hyperref}
\usepackage{caption}
\usepackage{comment}

\DeclareMathOperator{\E}{\mathbb{E}}

\newcommand\new[1]{\textcolor{black}{#1}}
\newcommand\neww[1]{\textcolor{black}{#1}}

\usepackage{graphicx}
\usepackage{color}
\usepackage{soul}
\usepackage{dblfloatfix}
\usepackage{cite}
\usepackage{hyperref}
\usepackage{xcolor}

\hypersetup{%
  colorlinks=true,% hyperlinks will be coloured
  linkcolor=blue,% hyperlink text will be green
  linkbordercolor=blue,% hyperlink border will be red
}

\title{\LARGE \bf
Deep Generative Modeling of LiDAR Data
}
\author{Lucas Caccia$^{1,2}$, Herke van Hoof$^{1,4}$, Aaron Courville$^{2,3}$, Joelle Pineau$^{1,2,3}$
\thanks{$^1$ MILA, McGill University}%
\thanks{$^2$ MILA, Universit\'e de Montr\'eal}%
\thanks{$^3$ CIFAR Fellow}%
\thanks{$^4$ University of Amsterdam}
}

\begin{document}

\maketitle
\thispagestyle{empty}
\pagestyle{empty}

%%%%%%%%%%%%%%%%%%%%%%%%%%%%%%%%%%%%%%%%%%%%%%%%%%%%%%%%%%%%%%%%%%%%%%%%%%%%%%%%
\begin{abstract}

Building models capable of generating structured output is a key challenge for AI and robotics. While generative models have been explored on many types of data, little work has been done on synthesizing lidar scans, which play a key role in robot mapping and localization. In this work, we show that one can adapt deep generative models for this task by unravelling lidar scans into a 2D point map. Our approach can generate high quality samples, while simultaneously learning a meaningful latent representation of the data. \neww{We demonstrate significant improvements against state-of-the-art point cloud generation methods.} Furthermore, we propose a novel data representation that augments the 2D signal with absolute positional information. We show that this helps robustness to noisy and imputed input; the learned model can recover the underlying lidar scan from seemingly uninformative data.
\end{abstract}

%%%%%%%%%%%%%%%%%%%%%%%%%%%%%%%%%%%%%%%%%%%%%%%%%%%%%%%%%%%%%%%%%%%%%%%%%%%%%%%%
\section{INTRODUCTION}

One of the main challenges in mobile robotics is the development of systems capable of fully understanding their environment. This non-trivial task becomes even more complex when sensor data is noisy or missing. An intelligent system that \neww{can replicate} the data generation process is much better equipped to tackle inconsistency in its sensor data. 
There is significant potential gain in having autonomous robots equipped with data generation capabilities which can be leveraged for reconstruction, compression, or prediction of the data stream.

In autonomous driving, information from the environment is captured from sensors mounted on the vehicle, such as cameras, radars, and lidars. While a significant amount of research has been done on generating RGB images, relatively little work has focused on generating lidar data. These scans, represented as an array of three dimensional coordinates, give an explicit topography of the vehicle's surroundings, potentially leading to better obstacle avoidance, path planning, and inter-vehicle spatial awareness. 

To this end, we leverage recent advances in deep generative modeling, namely variational autoencoders (VAE) \cite{vae} and generative adversarial networks (GAN) \cite{gan}, to produce a generative model of lidar data. While the VAE and GAN approaches have different objectives, they can be used \new{in conjunction} with Convolutional Neural Networks (CNN) \cite{cnn} to extract local information from nearby sensor points.

Unlike some approaches for lidar processing, we do not convert the data to voxel grids \cite{voxelnet, engelcke2017vote3deep}. Instead, \new{we build off existing work \cite{li2016vehicle} which projects the lidar scan into a 2D spherical point map}. \new{We show that this representation is fully compatible with deep architectures previously designed for image generation. Moreover, we investigate the robustness of this approach to missing or noisy data, a crucial property for real world applications. We propose a simple, yet effective way to improve the model's performance when the input is degraded. Our approach consists of augmenting the 2D map with absolute positional information, through extra  $(x,y,z)$ coordinate channels. We validate these claims through a variety of experiments on the KITTI \cite{geiger2013vision} dataset.}

\new{Our contributions are the following: 
\begin{itemize}
    \item We provide a fully unsupervised method for both \textit{conditional} and \textit{unconditional} lidar generation. 
    \item We establish an evaluation framework for lidar reconstruction, allowing the comparison of methods over a spectrum of different corruption mechanisms. 
    \item We propose a simple technique to help the model process noisy or missing data.
\end{itemize}
}

% \st{The main contribution of this paper is to provide a method for \textit{raw} lidar data generation that requires minimal human craftmanship. In other words, our work enables lidar point cloud generation in a fully unsupervised fashion. To the best or our knowledge, no previous attempts have been made to tackle this problem.  Experimental results using the standard KITTI dataset \cite{geiger2013vision} show that the method can synthesize lidar data that preserves essential structural properties of the scene.} 

\begin{figure}[!tbp]
\centering
  \includegraphics[width=.48\textwidth]{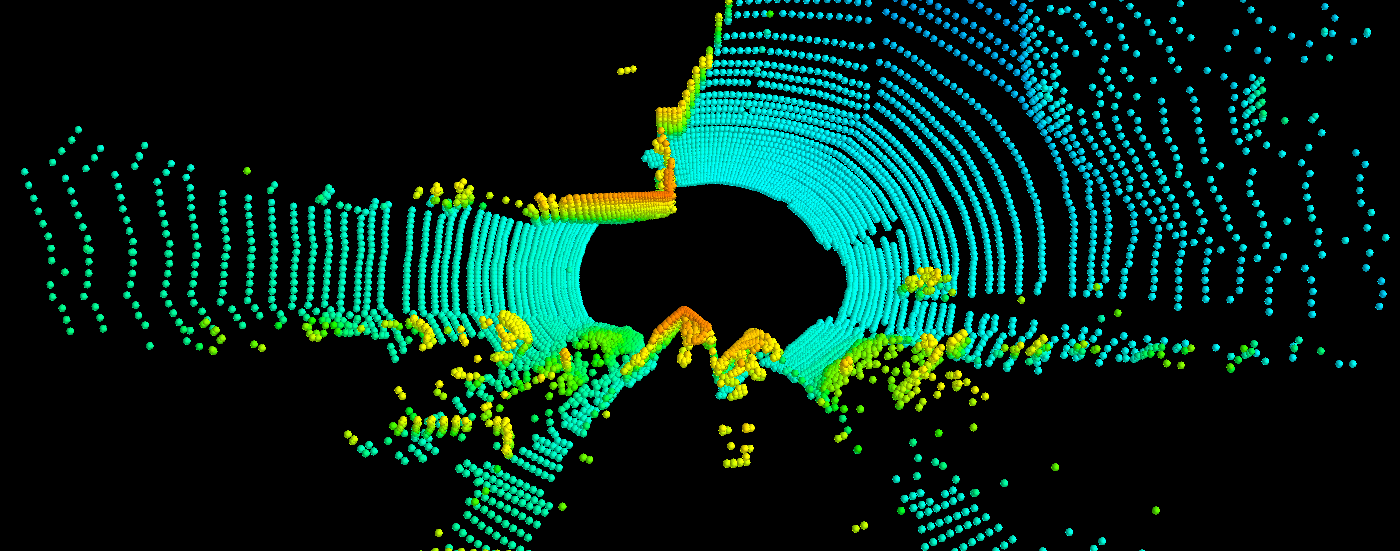}  \\
  \vskip 2px
  \includegraphics[width=.48\textwidth]{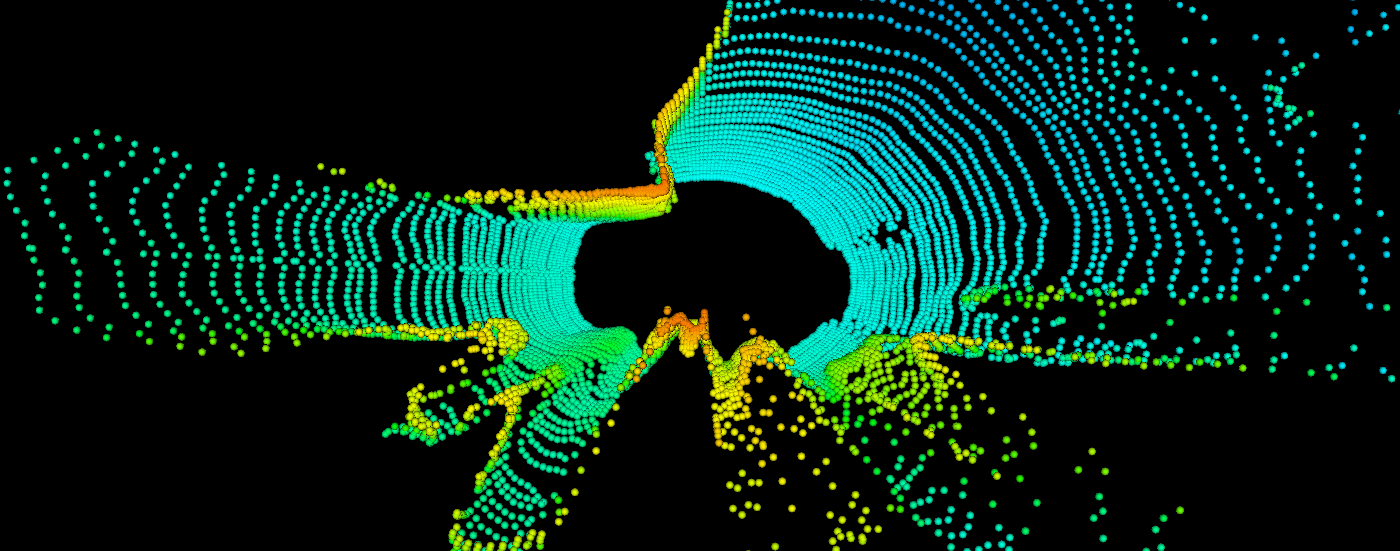}
    \vskip 2px
  \includegraphics[width=.48\textwidth]{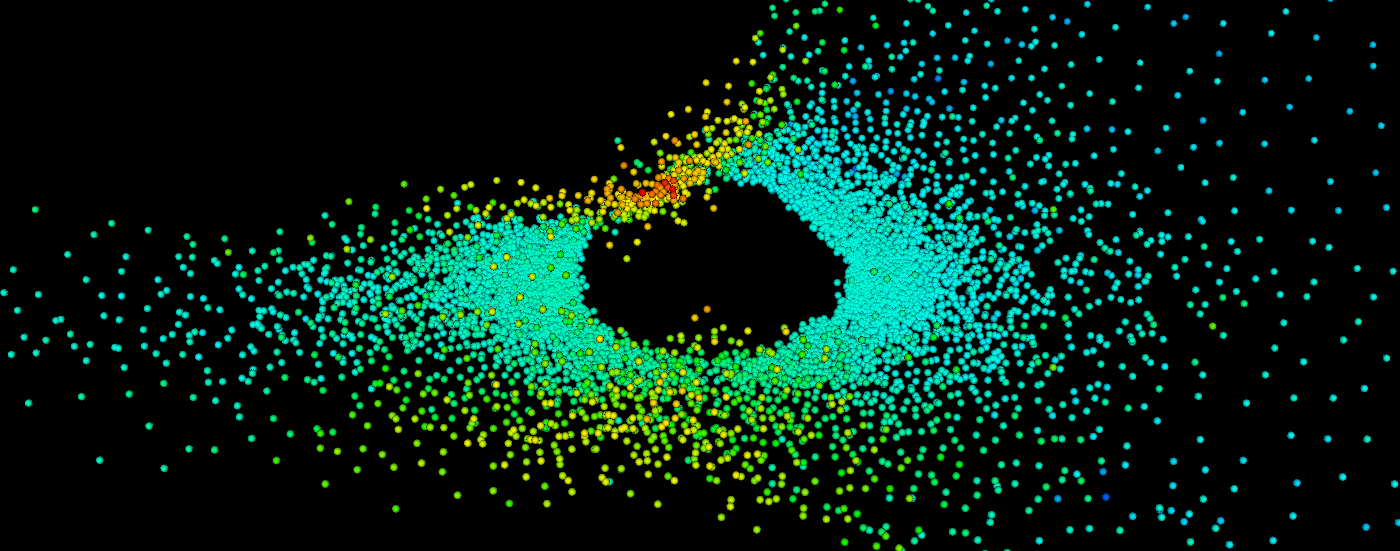}
% * <hercky.amazon@gmail.com> 2017-09-15T20:49:39.379Z:
% 
% I will recommend having a border between the images, as of now it gives the impression that this is a single image
% 
% ^.
  \caption{Best viewed in color. Top: real LiDAR sample from the test set. Middle: reconstruction from our proposed model. Bottom: reconstruction from the baseline model.}
\label{fig:recons}
\end{figure}

\section{Related work}
% TODO: put something here. 
% We first review how recent deep learning methods have been used for 

\subsection{Lidar processing using Deep Learning}
\label{bijection}
The majority of papers applying deep learning methods to lidar data present discriminative models to extract relevant information from the vehicle's environment. Dewan et al. ~\cite{dewan2017deep} propose a CNN for pointwise semantic segmentation to distinguish between static and moving obstacles.  Caltagirone et al. \cite{caltagirone2017fast} use a similar approach to perform pixel-wise classification for road detection. To leverage the full 3D structure of the input, Bo Li \cite{li20173d} uses 3D convolutions on a voxel grid for vehicle detection. However processing voxels is computationally heavy, and does not leverage the sparsity of LiDAR scans. \new{Engelcke et al. \cite{engelcke2017vote3deep} propose an efficient 3D convolutional layer to mitigate these issues.}

\new{Another popular approach \cite{li2016vehicle, velas2018cnn, vaquero2017deconvolutional, vaquero2018deep} to avoid using voxels relies on the inherent two-dimensional nature of lidars. It consists of a} bijective mapping from 3D point cloud to a 2D point map, where $(x,y,z)$ coordinates are encoded as azimuth and elevation angles measured from the origin. \new{This can also be seen as projecting the point cloud onto a 2D spherical plane}. Using such a bijection lies at the core of our proposed approach for generative modeling of lidar data.

% With this encoding, \cite{baidu-efficient-encoding} do efficient prediction of full 3D bounding boxes for cars. \cite{efficient-rep} show that this encoding allows for very fast ground segmentation for lidar data. We opt for a similar input format in our work.

\begin{figure*}
  \centering
  \includegraphics[width=\linewidth]{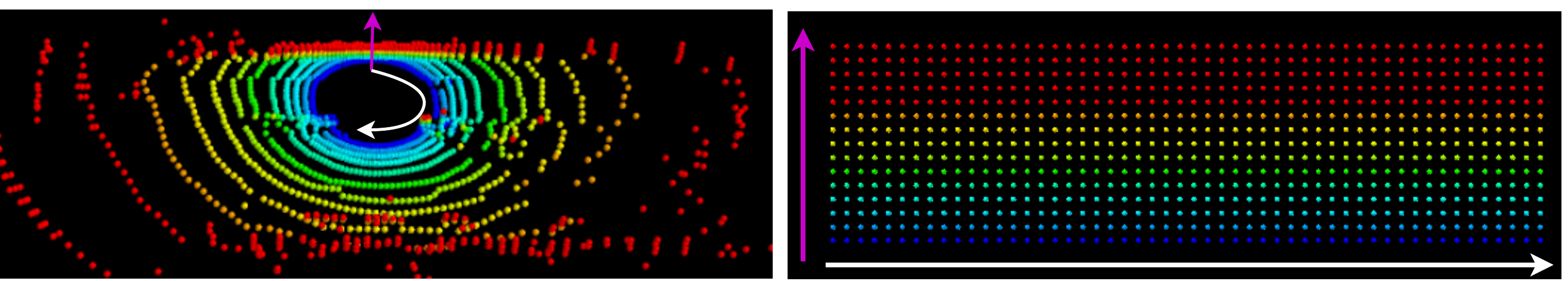}
  \caption{Best viewed in color. Our proposed ordering of points from 3D space (left) into a 2D grid (right). Points sampled from the same elevation angle share the same color. The ordering of every row is obtained by \textit{unrolling} points in increasing azimuth angle. The showed lidar was downsampled for visual purposes.}
  \label{fig:preprocessing}
\end{figure*}

\subsection{Grid-based lidar generation}
An alternative approach for generative modeling of lidar data is from Ondru´ska et al \cite{ondruska2016end}. They train a Recurrent Neural Network for semantic segmentation and convert their input to an occupancy grid. More relevant to our task, they train their network to also predict future occupancy grids, thereby creating a generative model for lidar data. Their approach differs from ours, \new{as the occupancy grid used assigns a constant area (400 cm$^2$) to every slot, whereas} we operate directly on \new{projected} coordinates. This not only reduces preprocessing time, but also allows us to efficiently represent data with non-uniform spatial density. We can therefore run our model at a much higher resolution, while remaining computationally efficient. 

\new{Concurrent with our work, Tomasello et al. \cite{tomasello2018dscnet} explore conditional lidar synthesis from RGB images. The authors use the same 2D spherical mapping proposed in \cite{li2016vehicle}. \neww{Our approach differs on several points. First, we do not require any RGB input for generation, which may not always be available (e.g. in poorly lit environments). Second, we explore ways to augment the lidar representation to increase robustness to corrupted data. Finally, we look at generative modeling of lidar data (compared to a deterministic mapping in their case).}} 

\subsection{Point Cloud Generation}
A recent line of work \cite{achlioptas2017representation, groueix2018atlasnet, yang2018foldingnet, fan2017point} considers the problem of generating point clouds as \textit{unordered sets} of $(x,y,z)$ coordinates. This approach does not define an ordering on the points, and must therefore be invariant to permutations. To achieve this, they use a variant of PointNet~\cite{qi2017pointnet} to encode a variable-length point cloud into a fixed-length representation. This latent vector is then decoded back to a point cloud, and the whole network is trained using permutation invariant losses such as the \textit{Earth-Mover's Distance} or the \textit{Chamfer Distance} \cite{fan2017point}. While these approaches work well for arbitrary point clouds, we show that they give suboptimal performance on lidar, as they do not leverage the known structure of the data.

\new{
\subsection{Improving representations through extra coordinate channels}
In this work, we propose to augment the 2D spherical signal with Cartesian coordinates. This can be seen as a generalization of the CoodConv solution \cite{liu2018intriguing}. The authors propose to add two channels to the image input, corresponding to the $(i,j)$ location of every pixel. They show that this enables networks to learn either complete translation invariance or varying degrees of translation dependence, leading to better performance on a variety of downstream tasks.
}

\section{Technical Background : Generative Modeling}

The underlying task of generative models is density estimation. Formally, we are given a set of $d$-dimensional \textit{i.i.d} samples $X = \{x_i \in \mathbb{R}^d\}^m_{i=1}$ from some unknown probability density function $p_{\textnormal{real}}$. Our objective is to learn a density $p_\theta$ where $\theta \in \mathcal{F}$ represents the parameters of our estimator and $\mathcal{F}$ a parametric family of models. Training is done by minimizing some distance $\mathcal{D}$ between $p_{\textnormal{real}}$ and $p_{\theta}$. The choice of both $\mathcal{D}$ and the training algorithm are the defining components of the density estimation procedure. Common choices for $\mathcal{D}$ are either $f$-divergences such as the Kullback-Liebler (KL) divergence, or Integral Probability Metrics (IPMs), such as the Wasserstein metric \cite{arjovsky2017wasserstein}. These similarity metrics between distributions often come with specific training algorithms, as we describe next.

\subsection{Maximum Likelihood Training}
Maximum likelihood estimation (MLE) aims to find model parameters that maximize the likelihood of $X$. Since samples are \textit{i.i.d}, the optimization criterion can be viewed as : 
\begin{equation}
    \max_{\theta \in \mathcal{F}} \E_{x \sim p_{\textnormal{real}}} \log(p_{\theta}(x)).
\end{equation}
It can be shown that training with the MLE criteria converges to a minimization of the KL-divergence as the sample size increases \cite{kolouri2018sliced}.
% MC bizarre ste phrase la. MLE ou min KL cest la mm affaire, non ?
From Eqn (1) we see that any model admitting a differentiable density $p_{\theta}(x)$ can be trained via backpropagation. Powerful generative models trained via MLE include Variational Autoencoders~\cite{vae} and autoregressive models \cite{oord2016pixel}. \new{In this work, we focus on the former, as the latter have slow sampling speed, limiting their potential use for real world applications}.

\subsubsection{Variational Autoencoders (VAE)}
The VAE \cite{vae} is a regularized version of the traditional autoencoder (AE). It consists of two parts: an inference network $\phi_{enc} \equiv q(z|x) $ that maps an input x to a posterior distribution of latent codes $z$, and a generative network $\psi_{dec} \equiv p(x|z)$ that aims to reconstruct the original input conditioned on the latent encoding. 
%MC pk cette notation ?? tu utilise p_\theta (x|z) deja et q_\phi plus loin ...
% In practice, both models are represented as neural networks. \\
By imposing a prior distribution $p(z)$ on latent codes, it enforces the distribution over $z$ to be smooth and well-behaved. This property enables proper sampling from the model via ancestral sampling from latent to input space. 
%\st{Without it, $\phi_{enc}$ would encode the inputs as single points by making the variance of $q(z|x)$ arbitrarily small\footnote{In this case the model collapses to a regular autoencoder.}}. \\ 
% MC q(z|x) souvent aura une variance presque nule mm sous un VAE. La KL va plus te pousser a remplir ton Gaussian bowl. Cest pas la mm chose.

The full objective of the VAE is then:
\begin{equation}
    \mathcal{L}(\theta;x) = \E_{q_{(z|x)}} \log p(x|z) -\textnormal{KL}(q(z|x) || p(z)) \leq \log p(x),
\end{equation}
which is a valid lower bound on the true likelihood, thereby making Variational Autoencoders valid generative models. For a more in depth analysis of VAEs, see \cite{doersch2016tutorial}.
% true log-likelihood

\subsection{Generative Adversarial Network (GAN)} The GAN~\cite{gan} formulates the density estimation problem as a minimax game between two opposing networks. 
% MC do gan really desnity estimate ?
The \textit{generator} $G(z)$ maps noise drawn from a prior distribution $p_{\textnormal{noise}}$ to the input space, aiming to fool its adversary, the \textit{discriminator} $D(x)$. 
% MC a tu deja definie le generator. p-e "A generator maps..."
The latter then tries to distinguish between real samples $x \sim p_{\textnormal{real}}$ and fake samples $x' \sim G(z)$. In practice, both models are represented as neural networks. %, and trained alternately via backpropagation. 
Formally, the objective is written as 
\begin{align} \label{gan_obj}
    \min_G \max_D \mathop{\mathbb{E}}_{x \sim p_{\textnormal{real}}} \log(D(x)) + \mathop{\mathbb{E}}_{z \sim p_{\textnormal{noise}}} \log(1 - D(G(z))).
\end{align}

GANs have shown the ability to produce more realistic samples \cite{karras2017progressive} than their MLE counterparts. However, the optimization process is notoriously difficult; stabilizing GAN training is still an open problem. In practice, GANs can also suffer from \textit{mode collapse} \cite{salimans2016improved}, which happens when the generator overlooks certain modes of the target distribution. 

\section{Proposed approach for lidar generation}
We next describe the proposed deep learning framework used for generative modeling of lidar scans.

\subsection{Data Representation}
Our approach relies heavily on 2D convolutions, therefore we start by converting a lidar scan containing $N$ $(x,y,z)$ coordinates into a 2D grid. We begin by clustering together points emitted from the same elevation angle into $H$ clusters. Second, for every cluster, we sort the points in increasing order of azimuth angle. In order to have a proper grid with a fixed amount of points per row, we divide the $\ang{360}$ plane into $W$ bins. \neww{This yields a $H \times W$ grid, where for each cell we store the average $(x,y,z)$ coordinate, such that we can store all the information in a $H \times W \times 3 $ tensor.} We note that the default ordering in most lidar scanners is the same as the one obtained after applying this preprocessing. Therefore, sorting is not required in practice, and the whole procedure can be executed in $\mathcal{O}(N)$. Figure \ref{fig:preprocessing} provides a visual representation of this mapping. \new{This procedure yields the same ordering of points as the projection discussed in \ref{bijection}. The latter would then return a grid of $H \times W \times 2$, where the $(x,y)$ channels are compressed as $d=\sqrt{x^2 + y^2}$. We will refer to the two representations above as \textit{Cartesian} and \textit{Polar} respectively. While this small change in representation seems innocuous, we show that when the input is noisy or incomplete, this compression can lead to suboptimal performance.}

\label{prepro}
%\new{\subsection{Preprocessing} If enough space add a few lines here}

%Since this mapping is implicitly in cylindrical coordinates (due to the ordering with respect to azimuth angle), we argue that the $(x,y,z)$ points should also be in cylindrical form: in this representation, the convolution operator becomes fully equivariant to rotations of the $(x,y)$ plane. Note that this is not the case when using $(x,y,z)$ values. We later show in the experimental section that this cylindrical mapping gives better samples, while showing similar quantitative performance. 

\subsection{Training Phase}

\subsubsection{VAEs} In practice, both encoder $\phi$ and decoder $\psi$ are represented as neural networks with parameters $\theta_{enc}$ and $\theta_{dec}$ respectively. 

Similar to a traditional AE, the training procedure first encodes the data $x$ into a latent representation $z = \phi(x; \theta_{enc})$. The variational aspect is introduced by interpreting $z$ not as a vector, but as parameters of a posterior distribution. In our work we choose a Gaussian prior and posterior, and therefore $z$ decomposes as $\mu_{x}, \sigma_{x}$.

We then sample from this distribution $\Tilde{z} \sim \mathcal{N}(\mu_x, \sigma_x)$ and pass it through the decoder to obtain $\tilde{x} = \psi(\tilde{z}; \theta_{dec})$.
Using the reparametrization trick \cite{vae}, the network is fully deterministic and differentiable w.r.t its parameters $\theta_{enc}$ and $\theta_{dec}$, which are updated via stochastic gradient descent (SGD).

\subsubsection{GANs} Training alternates between updates for the generator and discriminator, with parameters $\theta_{gen}$ and $\theta_{dis}$. Similarly to the VAE, samples are obtained by ancestral sampling from the prior through the generator. In the original GAN, the networks are updated according to Eqn. \ref{gan_obj}. In practice, we use the Relativistic Average GAN (RaGAN) objective \cite{jolicoeur2018relativistic}, which is easier to optimize. 
%Further details on the importance the RaGAN loss are provided in the appendix \ref{adv_details}.
Again, $\theta_{gen}$ and $\theta_{dis}$ are updated using SGD.
For a complete hyperparameter list, we refer the reader to our publicly available source code.\footnote{\url{https://www.github.com/pclucas14/lidar_generation}}

\subsection{Model Architecture} \label{model_arc}
Deep Convolutional GANs (DCGANs) \cite{dcgan} have shown great success in generating images. They use a symmetric architecture for the two networks: The generator consists of five transpose convolutions with stride two to upsample at each layer, and ReLU activations. The discriminator uses stride two convolutions to downsample the input, and Leaky ReLU activations. In both networks, Batch Normalization \cite{bn} is interleaved between convolution layers for easier optimization.
% MC batch norm is interleaved ?
We use this architecture for all our models: The VAE encoder setup is simply the first four layers of the discriminator, and the decoder's architecture replicates the DCGAN generator. 
 \new{We note that for both models, more sophisticated architectures \cite{zhang2018self, kingma2016improved} are fully compatible with our framework. We leave this line of exploration as future work.}

\section{Experiments}
This section provides a thorough analysis of the performance of our framework fulfilling a variety of tasks related to generative modeling.
First, we explore \textit{conditional} generation, where the model must compress and reconstruct a (potentially corrupted) lidar scan. We then look at \textit{unconditional} generation. In this setting, we are only interested in producing realistic samples, which are not explicitly tied to a real lidar cloud.

\subsection{Dataset}
We consider the point clouds available in the KITTI dataset \cite{geiger2013vision}. We use the train/validation/test set split proposed by \cite{lotter2016deep}, which yields 40 000, 80 and 700 samples for train, validation and test sets. We use the preprocessing described in section \ref{prepro} to get a $40 \times 256$ grid. For training we subsample from 10 Hz to 3 Hz since temporally adjacent frames are nearly identical.

\subsection{Baseline Models}
Since, to the best of our knowledge, no work has attempted generative modeling of raw lidar clouds, we compare to our method models that operate on arbitrary point clouds. We first choose AtlasNet \cite{groueix2018atlasnet}, which has shown strong modeling performance on the Shapenet \cite{chang2015shapenet} dataset. This network first encodes point clouds using a shared \textit{MLP} network that operates on each point \textit{individually}. A max-pooling operation is performed on the point axis to obtain a fixed-length global representation of the point cloud. In other words, the encoder treats each point independently of other points, without assuming an ordering on the set of coordinates. This makes the feature extraction process invariant to permutations of points. The decoder is given  the encoder output along with $(x,y)$ coordinates of a 2D-grid, and attempts to \textit{fold} this 2D-grid into a three-dimensional surface. The decoder also uses a \textit{MLP} network shared across all points.

Similar to AtlasNet, we compare our model with the one from Achlioptas et al \cite{achlioptas2017representation}. Only its decoder differs from AtlasNet; the model does not deform a 2D grid, but rather uses fully-connected layers to convert the latent vector into a point cloud, \new{making it less parameter efficient}. 

Both networks are trained end-to-end using the \textit{Chamfer Loss} \cite{fan2017point}, defined as 
\begin{equation}
d_{\textnormal{CH}} = \sum_{x \in S_1} \min_{y \in S_2} ||x - y||^2_2 + \sum_{y \in S_2} \min_{x \in S_1} || x - y ||^2_2,
\end{equation}
where $S_1$ and $S_2$ are two sets of $(x,y,z)$ coordinates. We note again that this loss is invariant to the ordering of the output points. For both autoencoders, we regularize their latent space using a Gaussian prior to get a valid generative model. 

% \subsection{Tasks}

 \begin{figure*}[!h]
     \centering
     \captionsetup{justification=centering}
     \includegraphics[width=.85\textwidth]{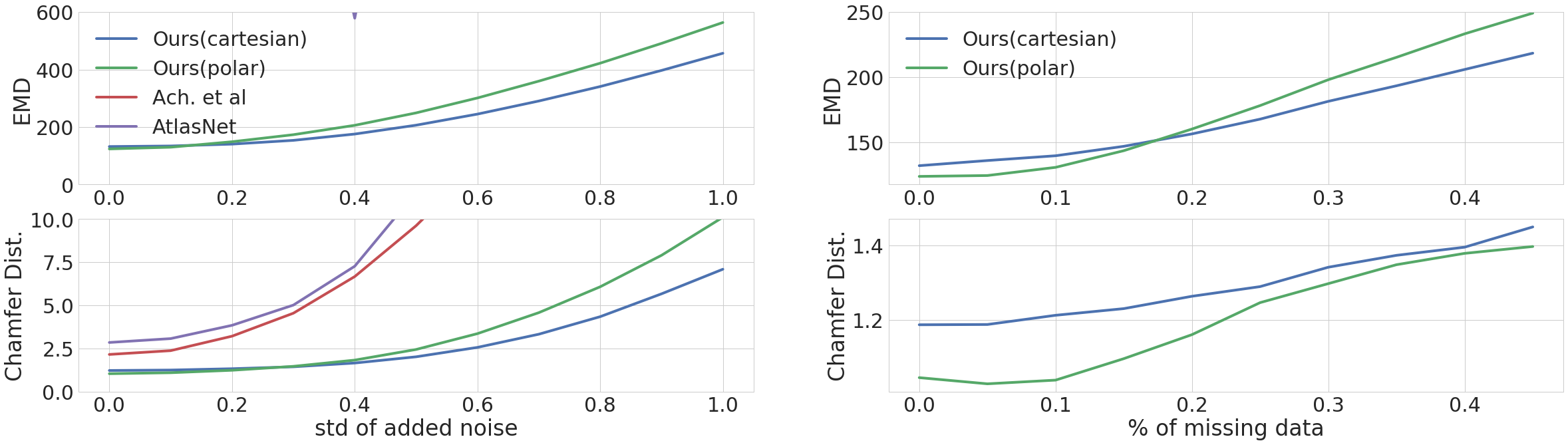}
     \caption{EMD and Chamfer Distance under varying levels of added noise (left) and missing data (right). We remove models with poor performance for clarity. For both metrics lower is better.
     % EMD and Chamfer Distance under different corruption mechanisms (lower is better). Left column shows varying levels of added noise. Right column displays different levels of missing data. We remove models with poor performance for clarity.
     }
     \label{fig:corrupted_graph}
 \end{figure*}

\subsection{Conditional Generation}
We proceed to test our approach in \new{a conditional generation task}. \new{In this setting, we do not evaluate the GAN, as this family of model -in their original formulation-} does not have an inference mechanism. 
\new{In other words, we consider four models: our approach, using either the Cartesian or the Polar representation, and the two baselines above. Since we are not sampling, but rather reconstructing an input, we consider both VAE and AE variants of every model, and report the best performing one.}

Formally, given a lidar cloud, we evaluate a model's ability to reconstruct it from a compressed encoding. More relevant to real word applications, we look at how robust the model's latent representation is to input perturbation. Specifically, we look at the two following corruption mechanisms:  
\begin{itemize}
    \item \textbf{Additive Noise} : we add Gaussian noise drawn from $\mathcal{N}(0, \sigma)$ to the $(x,y,z)$ coordinates of the lidar cloud. \new{For this process, we normalize each of the three dimension independently prior to noise addition}. We experiment with varying levels of $\sigma$.
    \item \textbf{Data Removal} : We remove random points from the input lidar scan. Specifically, the probability of removing a point is modeled as a Bernoulli distribution parametrized by $p$. We consider different values for $p$.
\end{itemize}

\subsection{Unconditional Generation}
For this section, we consider the GAN model introduced in section \ref{model_arc}. Our goal is to train a model that can produce realistic samples. \new{Having access to such a generator can lead to better simulator development, which are heavily used to train self-driving agents \cite{dosovitskiy2017carla}.} In this use case, an agent operating in an environment that lacks \textit{crispness} will likely result in poor skill transfer to real world navigation. Since the use of GANs has been shown to produce more realistic samples than MLE based models on images \cite{larsen2015autoencoding}, we hope to see similar results with our model in the case of LiDAR data. 

\textbf{Evaluation criteria}: Rigorous quantitative evaluation of samples produced by GANs and generative models is an open research question. GANs trained on images have been evaluated by the Inception Score \cite{salimans2016improved} and the Frechet Inception Distance (FID) \cite{heusel2017gans}. Since there exists no standardized metric for unconditional generation of lidar clouds, we rely on visual inspection of samples for quality assessment.

\subsubsection{Evaluation criteria}
%MC same here
To measure how close the reconstructed output is to the original point cloud, we use the \textit{Earth-Mover's Distance} \cite{fan2017point}. It is defined as 
\begin{equation}
d_{\textnormal{EMD}}(S_1, S_2) = \min_{\gamma:S_1 \longrightarrow S_2} \sum_{x \in S_1} || x - \gamma(x) ||_2 
\end{equation}where $\gamma$ is a bijection between the two sets.

The EMD gives the solution to the optimal transportation problem, which attempts to transform one point cloud into the other. Recent work \cite{achlioptas2017representation} has shown that this metric correlates well with human evaluation, \new{and does so better than the Chamfer Distance}. Moreover, the Earth Mover's Distance is sensitive to both global and local structure, and does not require points to be ordered. \new{Additionally, training and evaluating models on the same metric can result in models overfitting to this criterion, at the expense of sample quality \cite{wu2016google}. Nevertheless, we also provide results measured by the Chamfer Distance for completeness}. 

\new{
\subsubsection{Training Protocol} For every model considered, we perform the same hyperparameter search. We randomly select the learning rate, the latent dimension and the batch size from a predetermined set of values. This set of values is the same for all models to ensure fairness. This process is repeated for 10 different configurations, from which we choose the one obtaining the best performance on the validation set. We then proceed to evaluate this configuration on the test set according to the metrics described above. All models are trained end-to-end on the same dataset.
}

\section{Results}
In this section, we will first discuss results for conditional generation and subsequently evaluate results for unconditional generation of lidar images.

\begin{figure}[t]
\centering
  \includegraphics[width=.4\textwidth,trim={0cm 1cm 0 1cm},clip]{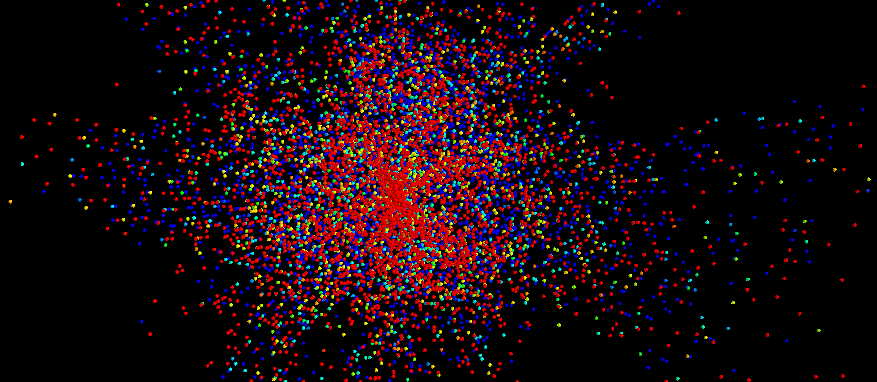}  \\
  \vskip 2px
  \includegraphics[width=.4\textwidth,trim={0cm 1cm 0 2.8cm},clip]{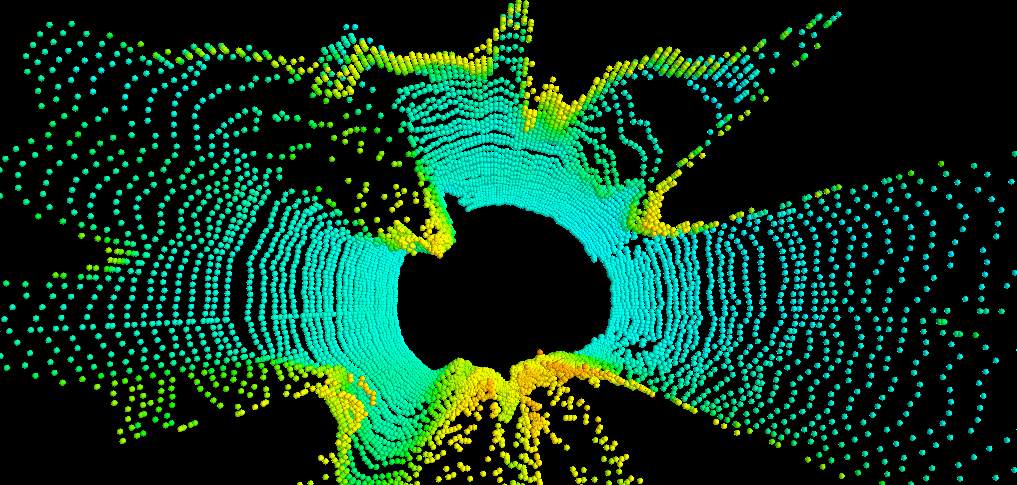} \\
    \vskip 2px
  \includegraphics[width=.4\textwidth,trim={0cm 1cm 0 1cm},clip]{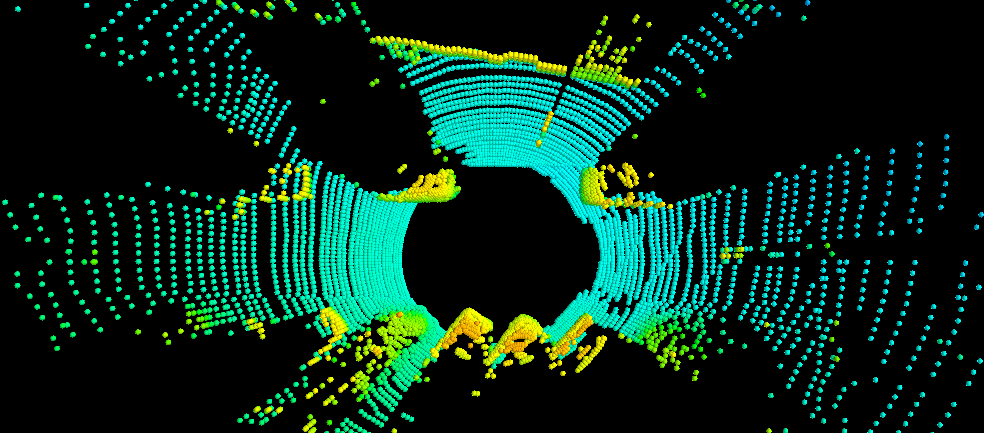}
  \caption{Top : corrupted lidar  from the test set, where we added noise drawn from $\mathcal{N}(0, 0.8)$ on the preprocessed scan. % This corresponds to a standard deviation of a, b, and c on the x, y  and z axis respectively. 
  Middle : reconstructed point cloud given corrupted input. Bottom : original lidar scan}
  % TODO: make this clearer.
 
  %MC emphasize how awesome this is
 \label{fig:noisy}
\end{figure}

\begin{figure*}[!h]
        \centering
        \includegraphics[width=.31\textwidth]{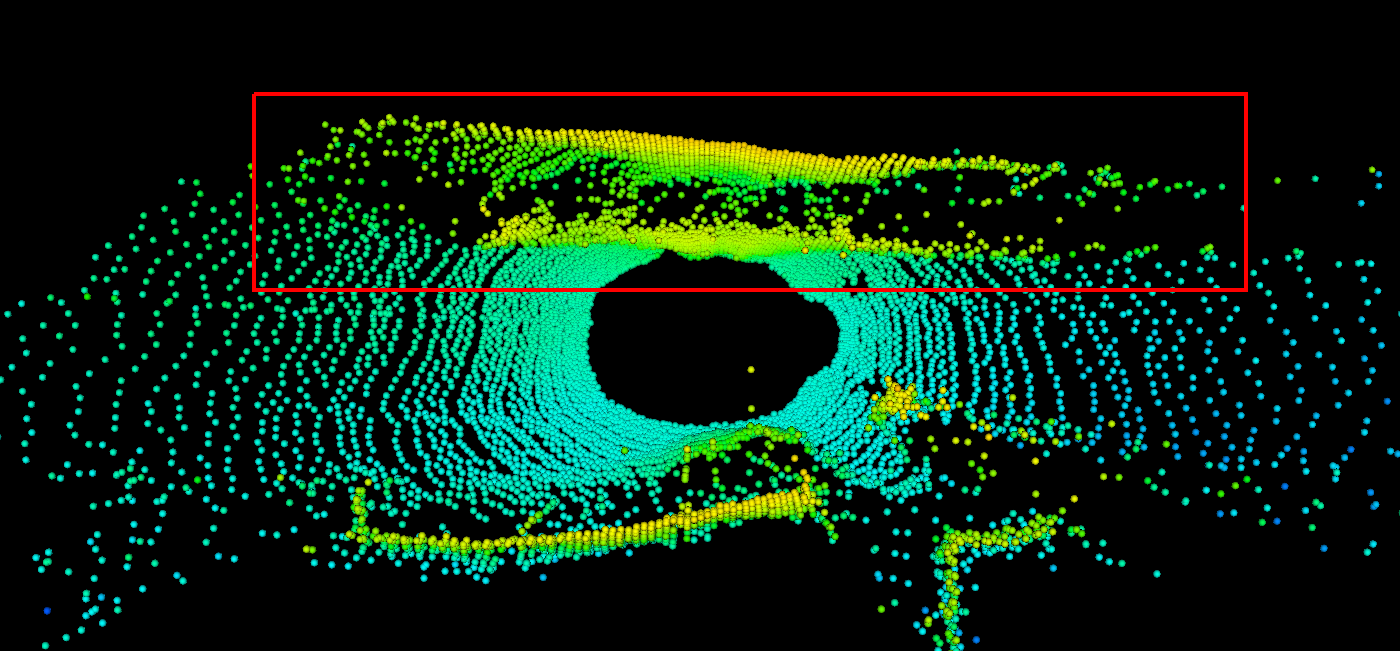} %\vskip 1mm
        \includegraphics[width=.31\textwidth]{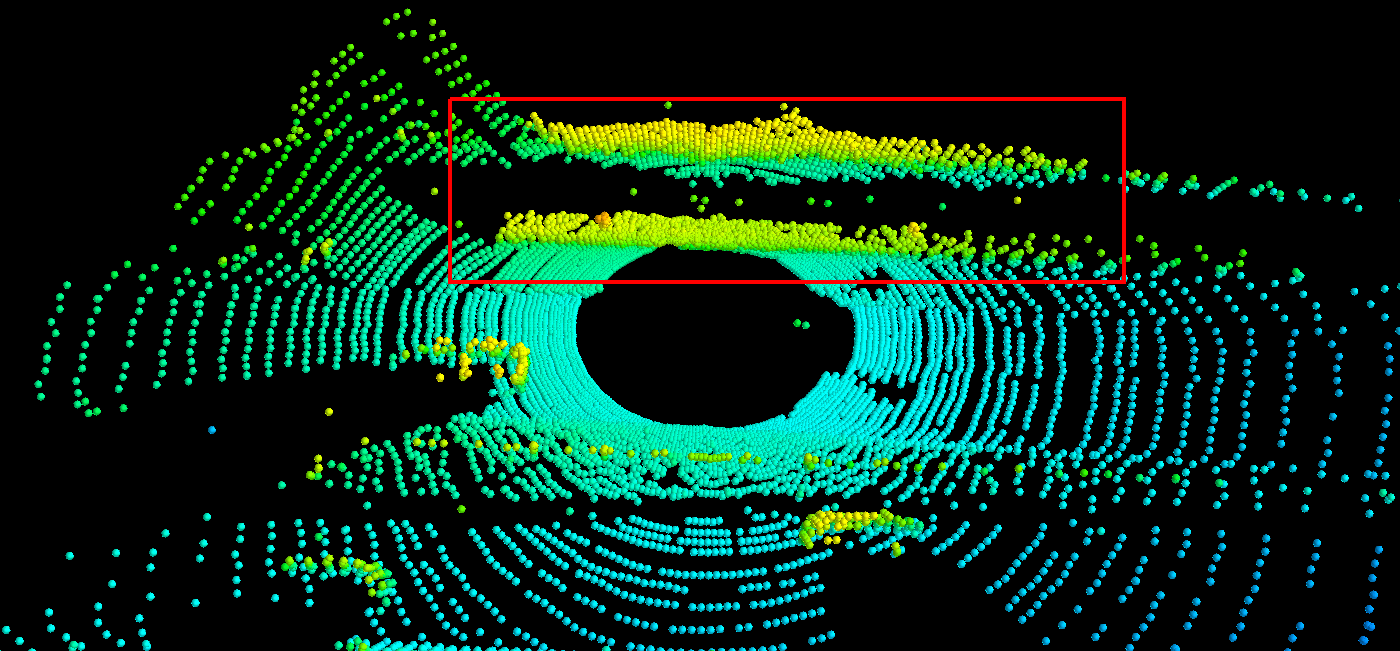} %\vskip 1mm
        \includegraphics[width=.31\textwidth]{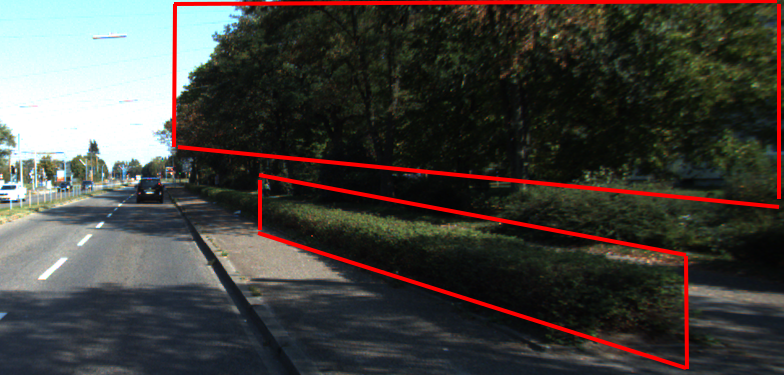} \vskip 2mm

        \includegraphics[width=.31\textwidth]{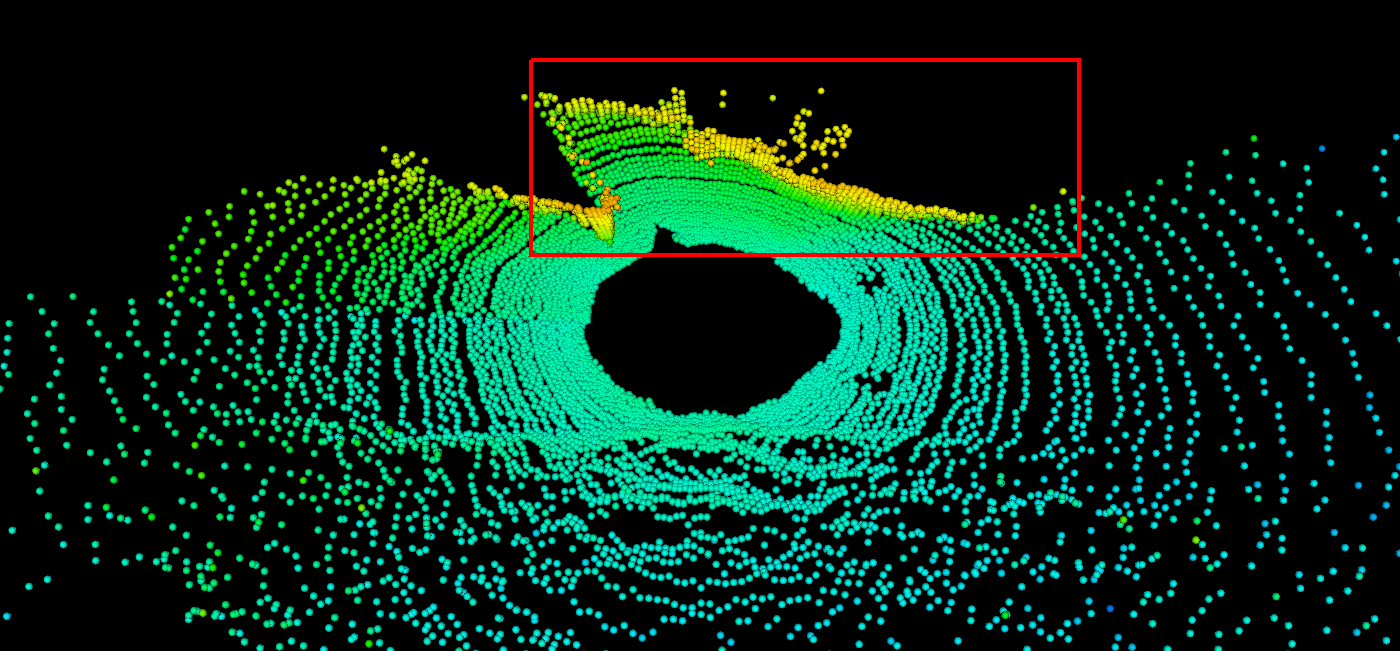} %\vskip 1mm
        \includegraphics[width=.31\textwidth]{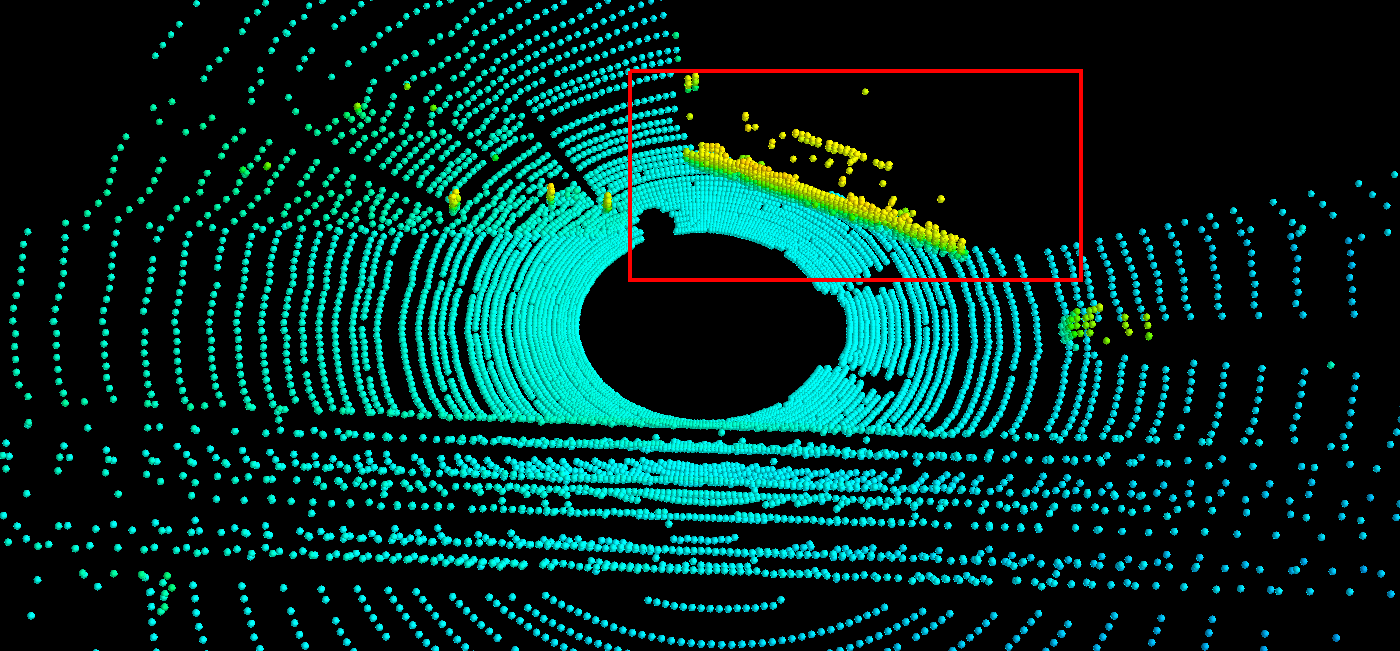} %\vskip 1mm
        \includegraphics[width=.31\textwidth]{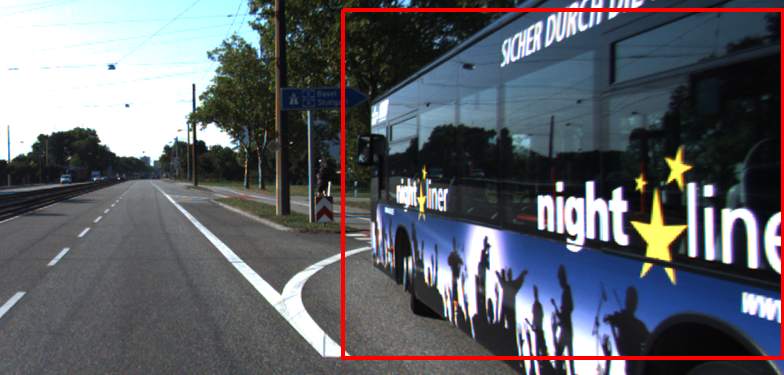} \vskip 2mm
        
        \includegraphics[width=.31\textwidth]{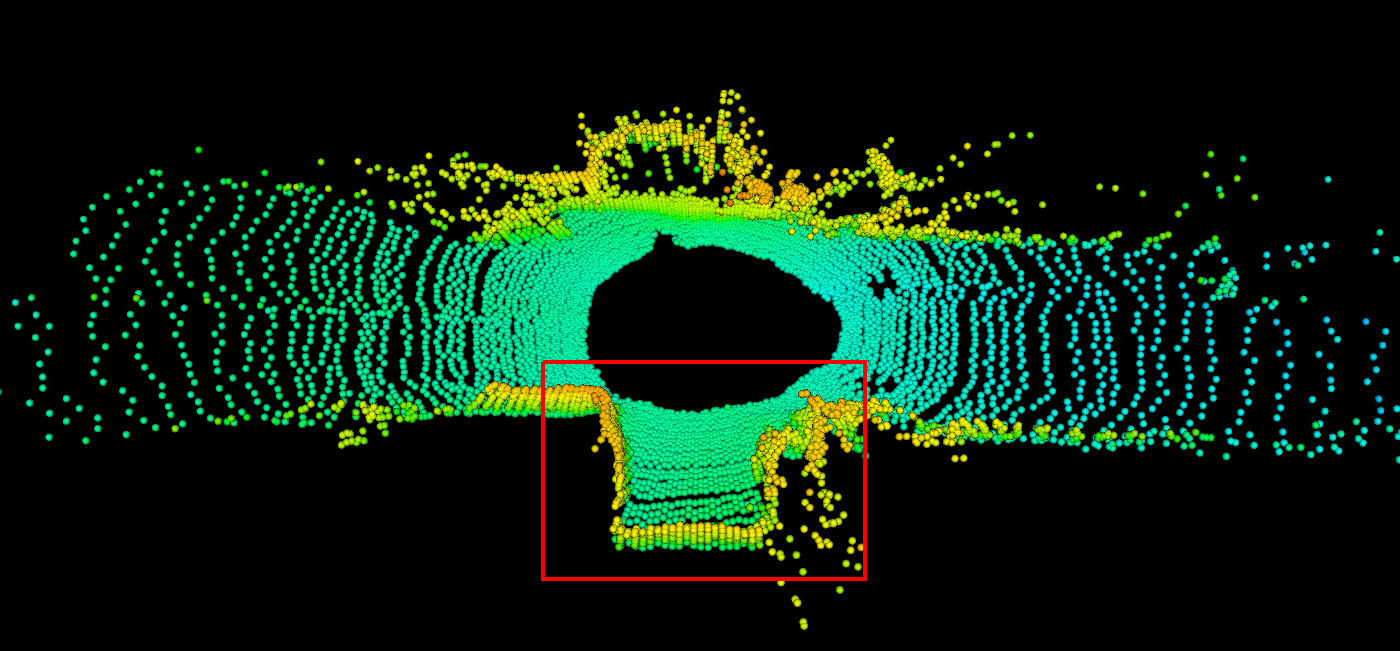} %\vskip 1mm
        \includegraphics[width=.31\textwidth]{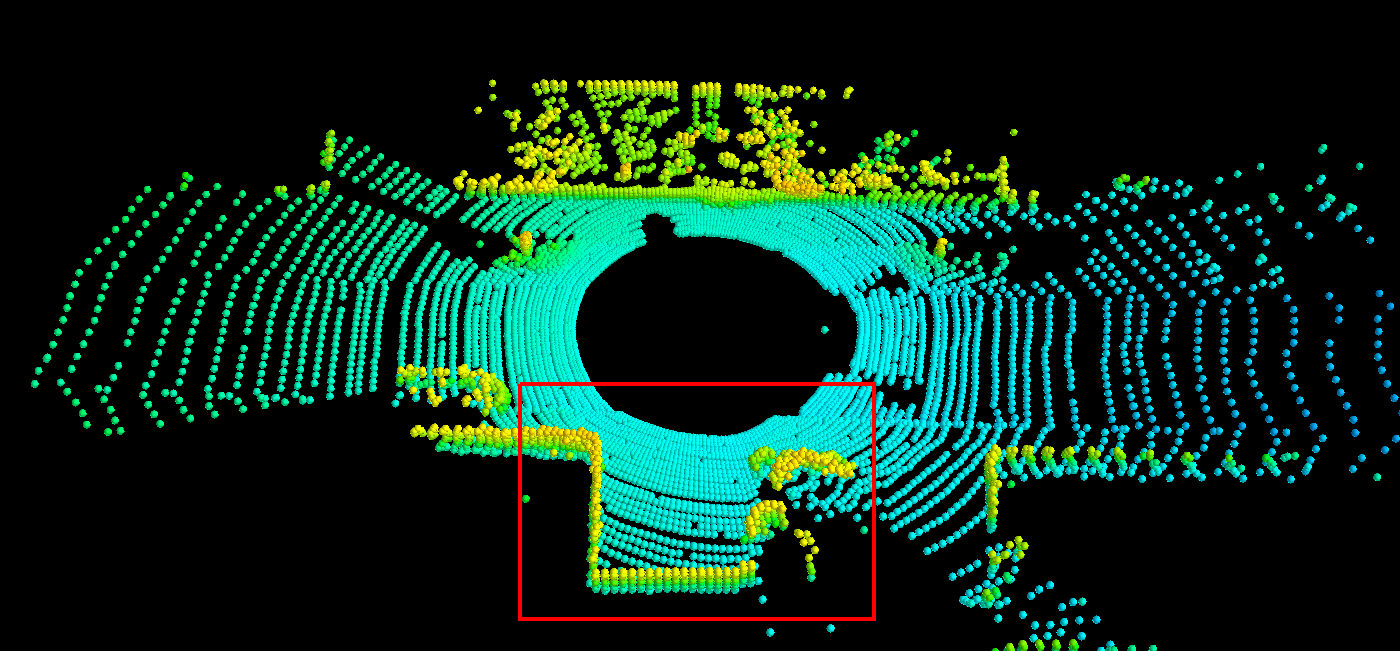} %\vskip 1mm
        \includegraphics[width=.31\textwidth]{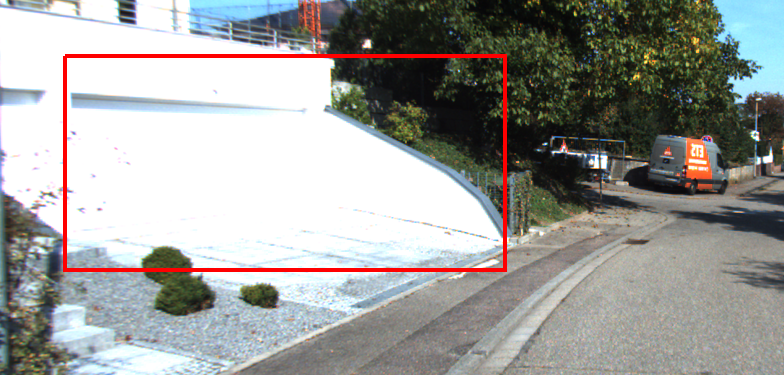} 
\caption{\new{We compare generated GAN samples (left) with their nearest neighbor in feature space (middle) from the test set. We show the corresponding RGB image (right). Regions of interest are highlighted in red. }}
\label{fig:imagination}
\end{figure*}

\subsection{Conditional}

In all conditional tasks, our proposed approach beats available baselines by a significant margin, both in terms of EMD, \new{Chamfer Distance} and visual inspection. 

\subsubsection{Reconstructing clean data}
\new{while the baseline models are able to reconstruct the global structure of the lidar scan, they are unable to recover the more fine grained detail of the input (see Fig.\ref{fig:recons}). This suggests that leveraging the known structure of the lidar plays a key role in obtaining high quality reconstructions.} Quantitative results are shown in Table \ref{tab:cond_quant}.

\begin{table}[!h]
\centering
\begin{tabular}[\textwidth]{ccc}
Model & EMD & Chamfer \\
\hline
\hline
Random  & 4331.9 & 253.6 \\
AtlasNet & 1571.2 & 2.85 \\
Ach. et al & 1103.1 & 2.16 \\
\hline
Ours(xyz) &  137.2  & 1.23 \\
\textbf{Ours(pol)} & \textbf{127.0}  & \textbf{1.04} \\
\end{tabular}
\caption{EMD and Chamfer distance measured on test set reconstructions (in both cases lower is better)}
\label{tab:cond_quant}
\end{table}

\subsubsection{Reconstructing corrupted data}

\new{
Next, we evaluate the proposed models on their ability to extract important information from corrupted lidar scans. As shown in Fig.   \ref{fig:noisy}, the proposed VAE correctly reconstructs the defining components of the original cloud, even if the given input is seemingly uninformative. We emphasize that \textbf{our model was not trained with such corrupted data}, therefore these results are quite surprising. Animations and additional reconstructions can be found \href{https://github.com/pclucas14/lidar_generation/\#reconstructions}{here}
}.

\new{Moreover, we observe that as soon as the input is moderately noisy, the proposed Cartesian representation yields better performance. As seen in Fig. \ref{fig:corrupted_graph}, this representation performs better than its Polar alternative over the majority of the graph. 
In addition, we observe a similar trend when points are randomly removed from the input, as shown in Fig. \ref{fig:corrupted_graph}; when more than 15\% of the points are missing, using $(x,y,z)$ coordinates performs favorably according to EMD. This result suggests that in this corruption regime, having access to absolute positional information provides a better signal to the model. Interesting future work would be to leverage the best of the two representations.}

% \TODO{A note on EMD vs Chamfer}

% \begin{figure*}
% \centering
% \begin{minipage}[b]{.4\textwidth}

% \includegraphics[width=.9\textwidth]{figs/out_noisy-1.png}

% \caption{EMD and Chamfer Distance for varying levels of added noise (lower is better).}
% \end{minipage}\qquad
% \begin{minipage}[b]{.4\textwidth}

% \includegraphics[width=.9\textwidth]{figs/out_missing.pdf}

% \caption{EMD and Chamfer Distance for varying levels of missing data (lower is better).}
% \end{minipage}
 %\end{figure*}

    We note that the suboptimal performance of the baselines is mainly due to two factors. First, since points are encoded independently, only information about the global structure is kept, and local fine-grained details are neglected. Second, the Chamfer Distance used for training assumes that the point cloud has a uniform density, which is not the case for lidar scans.

\subsection{Unconditional} 
\new{We perform a visual inspection of generated samples, located in the leftmost column of Figure \ref{fig:imagination} (more samples are available \href{https://github.com/pclucas14/lidar_generation/\#samples-from-gan-model}{here})}. We see that our model generates realistic samples. First, the scans have a well-defined global structure: an aerial view of the samples show points correctly aligned to model the structure of the road. Second, the samples share local characteristics of real data: the model correctly generates road obstacles, such as cars, or cyclists. This amounts to having locations with a dense aggregation of points, followed by a trailing area with almost no points, similar to the shadow of an object. Third, model respects the point density of the data, where the density is roughly inversely proportional to the distance from the origin. Lastly, our models show good sample diversity. % We refer to Figure \ref{fig:inter} for a specific analysis of samples, and to the code repository for a wider range of samples.

\new{
\subsubsection{What is the GAN generating?}
In order to better interpret samples from the unconditional generator, we try to match them to real data examples. We perform the following procedure: we encode every sample to a latent representation, given by the output of the third layer of our discriminator. We similarly encode random datapoints from the test set, and match the generated sample to the real datapoint yielding the smallest latent L2 loss. We show three examples of this matching in Figure \ref{fig:imagination}. In the first row, we see the model generating a two layer roadside to the right, consisting of a long shrub, followed by a line of trees. On the second row, we find a large tilted object to the right, which matches a bus turning right. Finally, 
on the last row we see a sharp enclosing, corresponding to a driveway leading to a garage door.  
}

\section{Discussion and Future Work}

In this work we introduced two generative models for \textit{raw} lidars, a GAN and a VAE. We have shown that the proposed adversarial network can generate highly realistic data, and captures both local and global features of real lidar scans. The LiDAR-VAE successfully encodes and reconstructs lidar samples, and is highly robust to missing or inputed data. We demonstrate that when adding enough noise to render the scan uninformative to the human eye, the proposed VAE still extracts relevant information and generates the missing data. \new{Our work in deep generative modeling of lidar enables concrete advancements in real life applications; the former model can help reduce the discrepancy between synthetic and real lidars in driving simulators, while the latter can be leveraged in deployed vehicles for reconstruction, compression, or prediction of the data stream.} \newline
\new{Moreover, we proposed a simple way to encode absolute positional information in the lidar representation, and showed that this leads to better reconstructions when the input is noisy or incomplete. Interesting future work would be to see if this can also lead to improvements in standard lidar processing tasks.}

%Moreover, we show that the LiDAR-GAN learns a meaningful latent representation. Interesting future work would be \st{to leverage this representation for sequential data generation. In other words, one could aim to train a model such that the latent space interpolations would produce temporally adjacent frames} \new{to explore if this new representation can give quantitative improvements for discriminative tasks more commonly associated with lidars.} 

\bibliographystyle{IEEEtran}
\bibliography{IEEEexample}

\newpage

\end{document}